\documentclass[sigconf]{acmart}

\AtBeginDocument{%
  }

\usepackage{subfig}
\usepackage{graphicx}
\usepackage{animate}
\usepackage{tcolorbox}
\usepackage{tabularx}
\usepackage{multirow}
\usepackage{array}
\usepackage[normalem]{ulem}

% ...

\newcommand{\eg}[0]{\emph{e.g.,}}
\newcommand{\ie}[0]{\emph{i.e.,}}
\newcommand{\etal}[0]{\emph{et al}}

\usepackage{etoolbox}
\usepackage{bm}
\usepackage{amsthm}%

\providecommand{\MyMapTemplatePrefixc}[4]{\expandafter#1\csname#3#4\endcsname{#2{#4}}} 
\forcsvlist{\MyMapTemplatePrefixc {\def} {\mathcal}{c}} {A,B,C,D,E,F,G,H,I,J,K,L,M,N,O,P,Q,R,S,T,U,V,W,X,Y,Z}  

\providecommand{\MyMapTemplatePrefixtb}[5]{\expandafter#1\csname#4#5\endcsname{#2{#3{#5}}}} 
\forcsvlist{\MyMapTemplatePrefixtb {\def} {\tilde}{\mathbf}{t}} {A,B,C,D,E,F,G,H,I,J,K,L,M,N,O,P,Q,R,S,T,U,V,W,X,Y,Z}  
\forcsvlist{\MyMapTemplatePrefixtb {\def} {\tilde}{\mathbf}{t}} {0,1,a,b,c,d,e,f,g,h,i,j,k,l,m,n,o,p,q,r,s,u,v,w,x,y,z}  

\providecommand{\MyMapTemplateNoPrefix}[3]{\expandafter#1\csname#3\endcsname{#2{#3}}}
\forcsvlist{\MyMapTemplateNoPrefix {\def} {\mathbf} } {0,1,a,b,c,d,e, f, g, h, i, j, k, l, m, n, o, p, q, r, u, v, w, x, y, z} 
\forcsvlist{\MyMapTemplateNoPrefix {\def} {\mathbf} } {A,B,C,D,E,F,G,H,I,J,K,L,M,N,O,P,Q,R,S,T,U,V,W,X,Y,Z}  

\def\etal{\emph{et al.}}
\def\ie{\emph{i.e.}}
\def\eg{\emph{e.g.}}

\usepackage{booktabs} 

\newcommand\blfootnote[1]{%
  \begingroup
  \renewcommand\thefootnote{}\footnote{#1}%
  \addtocounter{footnote}{-1}%
  \endgroup
}

\usepackage{amsfonts}
\usepackage{algpseudocode}
\usepackage[table]{xcolor} 
\usepackage{subcaption}
\usepackage{graphicx}
\usepackage{algorithm}
\usepackage{amsmath}
\usepackage[x11names]{xcolor}

\definecolor{topblue1}{RGB}{33,113,181}  
\definecolor{topblue2}{RGB}{107,174,214} 
\definecolor{topblue3}{RGB}{189,215,231} 

\newcommand{\topone}[1]{\cellcolor{topblue1}{#1}}
\newcommand{\toptwo}[1]{\cellcolor{topblue2}{#1}}
\newcommand{\topthree}[1]{\cellcolor{topblue3}{#1}}

\setcopyright{acmlicensed}
\copyrightyear{2026}
\acmYear{2026}
\acmConference[MM '26]{Proceedings of the 34th ACM International Conference on Multimedia}{November 10–-14 , 2026}{Rio de Janeiro, Brazil}

\acmSubmissionID{2502}

\copyrightyear{2026}
\acmYear{2026}
\acmConference[MM '26]{Proceedings of the 34th ACM International Conference on Multimedia}{November 10--14, 2026}{Rio de Janeiro, Brazil}
\acmBooktitle{Proceedings of the 34th ACM International Conference on Multimedia (MM '26), November 10--14, 2026, Rio de Janeiro, Brazil}
\acmDOI{10.1145/3767308.3835342}
\acmISBN{979-8-4007-2213-4/2026/11}

\begin{document}

\title{Versatile Video Representation via Feed-Forward 2D Gaussian Splatting Tokenization}

\settopmatter{authorsperrow=3}


\author{Zhenghao Chen$^{\dag*}$}
\affiliation{%
  \institution{The University of Newcastle}
  \city{Newcastle}
  \country{Australia}
}

\author{Zicong Chen$^{\dag}$}
\affiliation{%
  \institution{Beihang University}
  \city{Beijing}
  \country{China}
}

\author{Lei Liu}
\affiliation{%
  \institution{The University of Hong Kong}
  \city{Hong Kong SAR}
  \country{China}
}

\author{Yiming Wu}
\affiliation{%
  \institution{The University of Hong Kong}
  \city{Hong Kong SAR}
  \country{China}
}

\author{Dong Xu$^*$}
\affiliation{%
  \institution{The University of Hong Kong}
  \city{Hong Kong SAR}
  \country{China}
}

\renewcommand{\shortauthors}{Zhenghao Chen et al.}
\begin{abstract}
Recent video representation methods that rely on fixed-grid, patch-wise tokenization often exhibit limited versatility.
Spatially, uniformly allocating a fixed number of tokens often leads to over-encoding in low-information regions.
Temporally, reducing redundancy remains challenging without explicitly distinguishing between static and dynamic content.
In this work, we introduce the \textbf{\underline{G}}aussian \textbf{\underline{V}}ideo \textbf{\underline{T}}ransformer (\textbf{GVT}), a versatile video representation framework built on a feed-forward 2D Gaussian Splatting (2DGS) tokenization scheme. We first extract latent rigid features from a video clip and represent them with a set of 2D Gaussians generated by our proposed Spatio-Temporal Gaussian Embedding (STGE) mechanism in a feed-forward manner. Such 2D Gaussians not only enhance spatial adaptability by assigning higher (resp., lower) rendering weights to regions with higher (resp., lower) information content during rasterization, but also improve generalization by avoiding per-video optimization.
To enhance the temporal versatility, we introduce a Gaussian Set Partitioning (GSP) strategy that separates the 2D Gaussians into static and dynamic sets, which explicitly model static content shared across different time-steps and dynamic content specific to each time-step, enabling a compact representation.
We evaluate GVT across four tasks: video reconstruction, video action recognition, video compression, and video generation, on the UCF101, Kinetics, and DAVIS datasets. The results demonstrate state-of-the-art reconstruction and compression performance, improved action recognition, and video generation performance comparable to the baseline MAGVIT-v2.
\end{abstract}






\maketitle

\section{Introduction}
\label{sec:intro}

\begin{figure}[t!]
    \centering
    \includegraphics[width=0.48\textwidth]{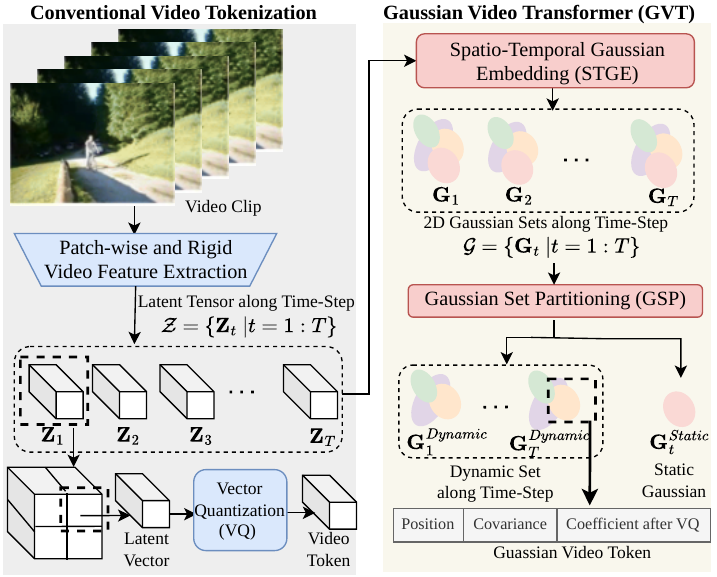}
    \caption{\textbf{Left:} rigidly designed conventional video tokenizer and \textbf{Right:} our proposed Gaussian Video Transformer (GVT).}
    \label{fig:overview}
\end{figure}

Video processing has become a core task in the digital era, underpinning a wide spectrum of multimedia applications~\cite{chen2023neural, chen2024group, Chen_2022_CVPR, hu2020improving, LIU2026114421, jiang2026neural}. 
To effectively support diverse video processing tasks, it is crucial to obtain compact yet expressive feature representations of videos (\ie, video representations). 
\blfootnote{$^\dag$The first two authors contributed equally. \\ 
$^*$Zhenghao Chen and Dong Xu are the Corresponding Authors. \\
\textit{E-mail: zhenghao.chen@newcastle.edu.au, dongxu@hku.hk}}
With the success of deep learning, particularly Visual Transformers~\cite{he2022masked, dosovitskiy2021vit}, large-scale video processing has entered a new era. These foundation models typically operate on sequences of visual tokens derived from raw image pixels, which is commonly referred to as video tokenization procedure.
An effective video tokenization method (\ie, tokenizer) can substantially reduce spatial and temporal redundancy~\cite{yulanguage}, improve representational efficiency, and ultimately boost the accuracy and scalability of video foundation models. For example, the recent video tokenizer MAGVIT-v2~\cite{yulanguage} achieves state-of-the-art reconstruction quality with compact representations while also demonstrating strong performance across a range of video processing tasks, including compression, understanding, and generation.

However, existing tokenizers exhibit limited versatility due to their rigid design. From a spatial perspective, they rely on fixed token grids without content-aware sampling, causing uniform or low-detail regions to be tokenized as densely as high-detail ones~\cite{yoa2025imagepiece, chen2022exploiting}. This results in inefficient representations, with a substantial proportion of tokens expended on uninformative regions. From a temporal perspective, most video tokenizers process time-step uniformly, without explicitly differentiating static content (\eg, background) from dynamic content (\eg, moving objects). Consequently, prolonged static segments receive comparable token budgets to rapidly varying segments, inducing redundant tokens for unchanging content and inflating the overall token count~\cite{yuframe}.

To enhance the versatility of video tokenization, we propose the \textit{Gaussian Video Transformer (GVT)}, which rasterizes video tokens via feed-forward 2D Gaussian Splatting (2DGS) to explicitly model parameters such as position, covariance, and coefficient. This design is inspired by recent advances~\cite{zhang2024gaussianimage, dong2025gaussiantoken, Tai_2025_CVPR, zhang2025gpstoken}, which 3D Gaussian Splatting (3DGS) to 2D image representations using radiance field techniques. By jointly leveraging these explicit and adaptive attributes, the resulting 2D Gaussians can efficiently capture local spatial structure, providing fine-grained modeling of regions with varying levels of detail and thereby improving the versatility of the learned representations.

On the other hand, effectively incorporating 2DGS into video remains a non-trivial challenge.
Recent works~\cite{Lee_2025_CVPR, bond2025gaussianvideo} directly optimize on individual videos tailored to specific per-video content, which requires additional optimization time and limits their generalizability.
To overcome this, we propose a \textit{Spatio-Temporal Gaussian Embedding (STGE)}, which generates 2D Gaussians in a feed-forward manner, inspired by recent advances in feed-forward 2DGS research~\cite{dong2025gaussiantoken}.
Specifically, we first apply a conventional video feature extraction to transform an input video clip into a sequence of rigid latent tensors across time-steps $T$.
STGE then projects these tensors into a set of 2DGSs $\cG = \{\G_t\}$ by leveraging two key modules: deformable spatio-temporal fusion (DSTF) and spatio-temporal attention (STA).
This design not only adaptively captures fine-grained spatial details within each time-step through explicit parameters, but also supports efficient single-pass inference, enabling scalable learning and generalization across diverse video content.

Although rasterizing independent 2D Gaussian sets at each time-step can intuitively encode video content, it primarily focuses on modeling dynamic transformations while neglecting those static components. As a result, the static content (\eg, background) is redundantly recomputed or duplicated across all frames, leading to unnecessary computational overhead and increased memory usage.
To mitigate this redundancy, we introduce a \textit{Gaussian Set Partitioning (GSP)} strategy that decomposes the 2DGS set $\cG$ into the dynamic and static Gaussians. The dynamic Gaussians model temporally varying content, while the static set captures stationary regions shared across all time steps and is stored only once, ensuring compactness.
Specifically, we learn a binary mask to decouple 2D Gaussians from the initial time-step's 2DGSs $\G_1$, designating them as static Gaussians. They are then reused across subsequent time-steps $\{t \mid t > 1\}$, avoiding using redundant tokens and improving the temporal versatility.

Overall, we evaluate GVT integrated with these two strategies on the video reconstruction task across the UCF101~\cite{soomro2012dataset}, Kinetics~\cite{kay2017kinetics}, and DAVIS~\cite{pont20172017} datasets, where it achieves state-of-the-art performance, highlighting its strong representational capacity. To further demonstrate the versatility of GVT, we evaluate it on three additional representative video processing tasks, namely action recognition, compression, and generation, using the UVG and Kinetics datasets. Our GVT consistently achieves state-of-the-art compression performance and surpasses MAGVIT-v2 in both recognition and generation. Our contributions are summarized as follows:

\begin{itemize}
\item \textbf{STGE}: A projection module that maps compact latent video representations to 2D Gaussian Splatting in a feed-forward manner, enabling versatile tokenization.

\item \textbf{GSP}: A learnable strategy that partitions static and dynamic 2D Gaussians to reduce temporal redundancy, enhancing temporal versatility and compactness.

\item \textbf{GVT}: An end-to-end video tokenization framework that integrates STGE and GSP, achieving state-of-the-art performance in reconstruction and compression, as well as competitive results in action recognition and generation.
\end{itemize}

\begin{figure*}[t!]
    \centering
    \includegraphics[width=\textwidth]{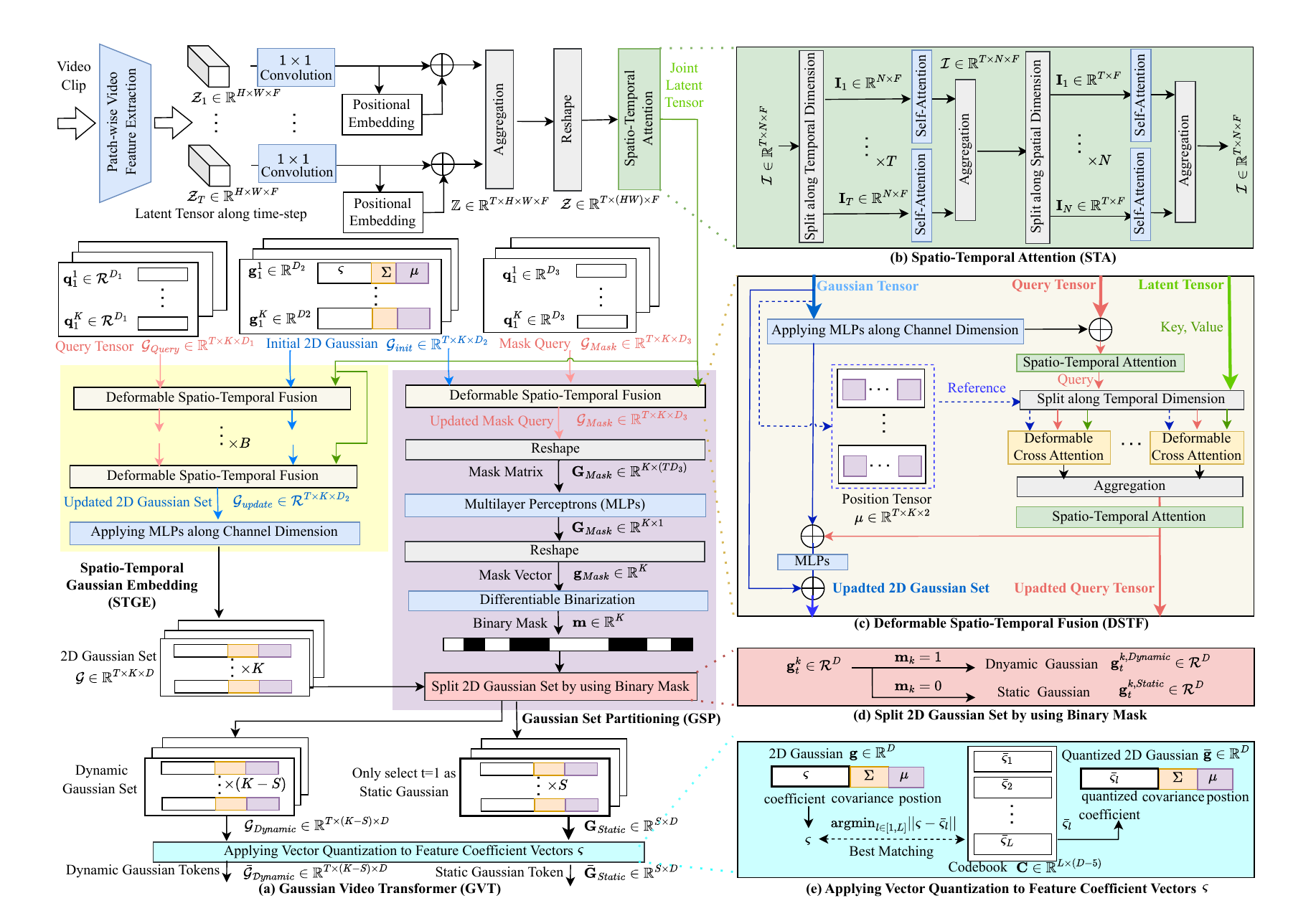} 
    \caption{(a) Overview of our GVT, built upon two core modules: Spatio-Temporal Gaussian Embedding (STGE) and Gaussian Set Partitioning (GSP). The others are: (b) Spatio-Temporal Attention (STA), (c) Deformable Spatio-Temporal Fusion (DSTF), (d) the binary-mask mechanism for Gaussian set splitting, and (e) the vector quantization (VQ) scheme for coefficient quantization.}
    \label{fig:framework}
\end{figure*}
\section{Related Works}
\subsection{Video Tokenizer}
Video tokenizer has emerged as a fundamental technique for converting videos into discrete representations.
As illustrated in Fig.\ref{fig:overview} (left), conventional video tokenizers typically partition a video into fixed-size, patch-wise vectors using Transformer-based encoders, followed by quantization into discrete tokens.
Early approaches such as TATS~\cite{ge2022long} and MAGVIT~\cite{yu2023magvit} adopt vector quantization (VQ) strategy based on a learnable codebook~\cite{esser2021taming}.

Recent studies~\cite{wang2024omnitokenizer, yulanguage, tan2024sweettok, wang2025vidtwin, zhaoimage} extended such VQ-based frameworks by introducing more powerful feature extractors and improved quantization strategies.
For example, MAGVIT-v2~\cite{yulanguage} integrates a causal 3D CNN with a shared vocabulary across images and videos, achieving superior performance compared to video diffusion models on standard benchmarks.
LARP~\cite{wanglarp} employs learned global queries for holistic video context modeling, and incorporates an autoregressive prior to structure latent representations, facilitating smoother and coherent generation.

The aforementioned methods primarily rely on fixed-grid, patch-wise tokenization strategies, which could limit their versatility in both spatial and temporal modeling.
In contrast, our GVT introduces a generative 2DGS strategy to produce Gaussian-Splatting-based video tokens that explicitly optimize both spatial structure and temporal dynamics.
This design significantly enhances the spatial and temporal versatility of the video tokenization process. To the best of our knowledge, this is the first work to incorporate Gaussian Splatting into the video tokenization research.

\subsection{Gaussian Splatting}
3DGS~\cite{kerbl20233d} has recently achieved remarkable success in 3D scene representation. Unlike Neural Radiance Fields (NeRFs)~\cite{mildenhall2021nerf}, which rely on implicit neural functions, 3DGS employs an explicit and versatile representation using Gaussian ellipsoids, which provides greater rendering efficiency.
Inspired by this, researchers have begun adapting 3DGS to 2D image~\cite{zhang2024gaussianimage, zhang2024image, Tai_2025_CVPR}.
For instance, Zhang~\etal~\cite{zhang2024gaussianimage} proposed GaussianImage, which leverages 2D Gaussian primitives for efficient image representation.

Subsequent work has extended 2DGS to video processing~\cite{liu2025d2gv, wang2025gsvc, Lee_2025_CVPR, bond2025gaussianvideo, sun2024splatter}.
These methods generally follow a high-level design inspired by deformable 3DGS~\cite{Wu_2024_CVPR}, in which the dynamic video content is disentangled into a canonical 2DGS representation and a corresponding 2D deformation field, which are then stored in light-weight data structure (\eg, MLPs).
For example, Lee~\etal~\cite{Lee_2025_CVPR} introduced a multi-plane spatio-temporal representation combined with a lightweight decoder to capture and reconstruct deformable information.
Nevertheless, such methods are not directly applicable as video tokenizers, as they rely on per-video optimization.
Consequently, they suffer from limited generalizability and cannot be applied to unseen videos without costly re-optimization.
Also, this per-sample retraining process is computationally expensive, making such methods impractical for real-time applications.

Our work is inspired by recent approaches that generate 2D or 3D Gaussian primitives in a feed-forward manner~\cite{huang2024gaussian, dong2025gaussiantoken, zhang2025gpstoken}.
However, they are primarily tailored to 3D scenes or static 2D images, and their applicability to 2D video remains largely unexplored.
To the best of our knowledge, the proposed GVT constitutes the first feed-forward 2DGS-based framework designed for video tokenization.

\section{Methodology}
\subsection{Preliminary}
\label{definition}
\subsubsection{Video Tokenization.} 
\label{def_vt}
As illustrated in Fig.~\ref{fig:overview} (left), most existing video tokenizers adopt a patch-wise and rigid tokenization scheme. Given a video clip $\mathbb{V} \in \mathbb{R}^{T' \times H' \times W' \times 3}$ consisting of $T'$ frames, where each frame $\cV \in \mathbb{R}^{H' \times W' \times 3}$ and $H'$, $W'$ denote the original height and width respectively, an encoder $\cE(\cdot)$ is typically employed to project $\mathbb{V}$ into a latent representation $\mathbb{Z} \in \mathbb{R}^{T \times H \times W \times F}$, where $T$, $H$, $W$, and $F$ are the temporal, spatial, and feature dimensions, typically reduced from the original video dimensions.
%
%
In this work, we follow~\cite{yu2023magvit, yulanguage} to disentangle the spatial and temporal dimensions and represent $\mathbb{Z} = \{\cZ_t | t=1:T\}$, where $\cZ_t \in \mathbb{R}^{H \times W \times F}$ is further represented into a set of $N = H \times W$ latent vectors $\z_t^i \in \mathbb{R}^{F}$, resulting in $\mathbb{Z}=\{\z_t^i |_{t=1:T}^{i=1:N}\}$. Here, $i \in [1, N]$ indexes the spatial position and $t \in [1, T]$ denotes the temporal time-step. The overall video embedding procedure is then denoted as:

\begin{equation}
\mathbb{Z} = \cE(\mathbb{V}), s.t., \mathbb{Z} =\{\z_t^i |_{t=1:T}^{i=1:N}\}.
\label{eq:encode}
\end{equation}

A quantizer $\cQ(\cdot)$ is then applied to these latent vectors. Most prior works directly adopt the VQ strategy~\cite{esser2021taming}, which maps each latent vector $\mathbf{z}_{t}^i$ to its nearest neighbor (\ie, codewords) in a learned codebook, denoted as $\bar{\z}_t^i = \cQ(\z_t^i)$. This process generates $\bar{\mathbb{Z}}=\{\bar{\z}_t^i |_{t=1:T}^{i=1:N}\}$, referred to as \textit{video tokens}.
A decoder $\cD(\cdot)$ subsequently decodes $\bar{\mathbb{Z}}$ back into the pixel space to obtain the reconstructed video $\bar{\mathbb{V}}$. The aim is to minimize $||\mathbb{V} - \bar{\mathbb{V}}||$.

\subsubsection{Video 2D Gaussian Splatting.} 
\label{def_2dgs}
As discussed, conventional video tokenizers exhibit limited versatility. In this work, instead of directly applying the VQ strategy on $\mathbb{Z}$ to obtain video tokens $\mathbb{\bar{Z}}$, we propose to project $\mathbb{Z}$ into a set of 2D Gaussians, where each 2D Gaussian $\g \in \mathbb{R}^D$ is defined by a 2D position $\mu \in \mathbb{R}^{2}$, a covariance matrix $\Sigma \in \mathbb{R}^{2 \times 2}$, and a feature coefficient vector $\varsigma \in \mathbb{R}^{D - 5}$. Here, to compute $\Sigma$, we follow~\cite{zhang2024gaussianimage} to factorize $\Sigma$ into a rotation matrix $\S$ and a scaling matrix $\R$ using 
$
\Sigma = (\R\S)(\R\S)^T, s.t., 
\mathbf{R} =
\begin{bmatrix}
cos(\theta) - sin(\theta) \\
sin(\theta) \:\:\:\: cos(\theta)
\end{bmatrix}, 
\mathbf{S} =
\begin{bmatrix}
s_1 \: 0 \\
0 \: s_2,
\end{bmatrix},$
where $\theta$ denotes the rotation angle, while $s_1$ and $s_2$ represent scaling factors along the principal axes. As a result, $\Sigma$ can be described using three variables: $\theta$, $s_1$, and $s_2$.
%
Following the temporal partitioning of video tokens, we evenly divide the Gaussian set  $\cG \in \mathbb{R}^{T \times K \times D}$ into $T$ temporal segments (\ie, $\cG = \{ \G_t | t = 1:T \}$), where each segment $\G_t \in \mathbb{R}^{K \times D}$ contains $K$ Gaussians $\g_t^k \in \mathbb{R}^{D}$ (\ie, $\G_t = \{ \g_t^k \mid k = 1 : K \}$) that represent the video token tensor $\bar{\cZ}_t$ at time step $t$. Therefore, the complete 2DGS set can be expressed as $\cG = \{\g_t^k |_{t = 1 : T}^{k = 1 : K} \}$. In this work, we will use such $\cG$ as the representation of the video tokens $\bar{\mathbb{Z}}$.

\subsection{Gaussian Video Transformer}
\label{GVT}
\subsubsection{Overview.} The framework is shown in Fig.~\ref{fig:framework} (a). Specifically, our GVT first transforms the input video into the latent representation $\mathbb{Z}$ (as defined in \textit{Sec.~\ref{def_vt}}) using a patch-wise video feature extraction approach with temporally causal 3D convolutions, following~\cite{yulanguage}.
Next, we project $\mathbb{Z}$ into a set of 2DGSs $\cG$ (see \textit{Sec.~\ref{def_2dgs}}) using the STGE module, resulting in a total of $T \times K$ 2D Gaussians.

To enhance temporal versatility, we further introduce GSP mechanism, which disentangles $\cG$ into:  1) a static Gaussian set $\G_{\text{Static}} \in \mathbb{R}^{S \times D},$ containing $S$ 2DGSs, and  
2) a dynamic Gaussian set $\cG_{\text{Dynamic}} \in \mathbb{R}^{T \times (K-S) \times D}$, containing $(K - S)$ 2D Gaussian primitives for each time-step $t$, resulting in the final 2DGS set 
${\cG} = {\G}_{\text{Static}} \cup {\cG}_{\text{Dynamic}}$. 
Finally, we quantize the set $\cG$ into $\bar{\cG}$ by applying the VQ strategy to the coefficient vectors $\varsigma$, mapping each to a discrete coefficient vector $\bar{\varsigma}$. The quantized set $\bar{\cG}$ is then used to rasterize the tokens $\bar{\mathbb{Z}}$, which we refer such $\bar{\cG}$ to as \textit{Gaussian video tokens}.

\subsubsection{Spatio-Temporal Gaussian Embedding.}
\label{gvt_stge}
STGE is designed to embed the latent tensor $\mathbb{Z}$ into a 2D Gaussian set $\cG$, drawing inspiration from recent feed-forward Gaussian Splatting methods~\cite{huang2024gaussian, dong2025gaussiantoken}. As shown in Fig.~\ref{fig:framework} (a), we first decompose the latent representation $\mathbb{Z}$ into a set of latent tensors $\mathbf{\cZ}_t$. For each $\cZ_t$, we apply a standard $1 \times 1$ convolutional layer followed by positional embeddings. The resulting features are then aggregated and reshaped into a joint latent tensor $\cZ \in \mathbb{R}^{T \times N \times F}, s.t., N= H\times W$, which consists of $N\times T$ updated latent vectors $\z_t^s$ (\ie, $\cZ=\{\z_t^i |_{t=1:T}^{i=1:N}\}$).
To better exploit spatio-temporal correlations, we refine the joint latent tensor $\cZ$ using a STA module (Fig.~\ref{fig:framework} (b)).

We also initialize an initial 2D Gaussian set  $\cG_{\text{init}} \in \mathbb{R}^{T \times K \times D_2}$
and a learnable query tensor $\cG_{\text{query}} \in \mathbb{R}^{T \times K \times D_1}.$
Here, $\cG_{\text{init}} = \{\g_t^k |_{t = 1 : T}^{k = 1 : K}\}$ consists of $K$ initial 2DGS $\g_t^k \in \mathbb{R}^{D_2}$ at each time step $t$. Each $\g_t^k$ is  parameterized by its position $\mu_t^k \in \mathbb{R}^{2}$
covariance $\Sigma_t^k \in \mathbb{R}^{3}$, and coefficient $\varsigma_t^k \in \mathbb{R}^{D_1 - 5}$ (see 2DGS parameterization in \textit{Sec.~\ref{def_2dgs}}).

To map the rigid latent tensor $\cZ$ into the 2DGS space, we introduce the DSTF module (Fig.~\ref{fig:framework} (c)). This module leverages the learnable queries $\cG_{\text{query}}$ to project information from each latent vector $\mathbf{z}_{t}^i$, located at position $\mu_t^i$, into the most relevant Gaussian $\g_t^k$ at position $\mu_t^k$ within the same time step $t$.
After $B$ iterations of fusion using the DSTF module, we obtain the updated 2D Gaussian set $\cG_{\text{update}} \in \mathbb{R}^{T \times K \times D_2}$. This set is then passed through a MLP network to generate the final 2D Gaussian set $\cG$, as defined in \textit{Sec.~\ref{def_2dgs}}. 
The overall procedure can be denoted as

\begin{equation}
\cG = \text{STGE}(\mathbb{Z}), s.t., \cG =\{\g_t^k |_{t=1:T}^{i=1:K}\}.
\label{eq:stge}
\end{equation}

\subsubsection{Gaussian Set Partitioning.} 
\label{gvt_gsp}
Though we can use $\cG$ with $K \times T$ 2D Gaussians to represent, the temporal redundancy across time steps remains unexploited. In particular, some 2D Gaussian primitives may correspond to static content, such as background regions, whose parameters do not need to be updated at each time step $t$. Inspired by dynamic 3DGS methods~\cite{lee2024fully, wu2025swift4d}, we define these Gaussian primitives as the static Gaussian set $\G_{\text{static}}$ and propose a GSP method to disentangle $\G_{\text{static}}$ from the full set $\cG$.

Here, we re-use the initial 2D Gaussian Set $\cG_{init}$ and the produced joint latent tensor $\cZ$ used at STGE stage (See \textit{Sec.~\ref{gvt_stge}}), and we intialize a mask query tensor $\cG_{Mask} \in \mathbb{R}^{T\times K \times D_3}$. Similar to the STGE stage, we adopt another DSFT module to generate an updated mask matrix tensor  $\cG_{Mask} \in \mathbb{R}^{T\times K \times D_3}$, which is further reshaped into a mask matrix $\cG_{Mask} \in \mathbb{R}^{K \times (T D_3)}$. We feed this matrix into a standard MLPs network and reshape the results into a mask vector $\g_{Mask} \in \mathbb{R}^{K}$. Last, we apply a differentiable binarization approach using the Straight-Through Estimator (STE)~\cite{yinunderstanding} to convert $\g_{\text{Mask}}$ into a binary mask $\m \in \mathbb{R}^{K}$ of $K$ dimensions.

Finally, we apply the produced binary mask $\m$ to partition the Gaussian set $\cG$, as illustrated in Fig.~\ref{fig:framework} (d). For each time step $t$ and Gaussian index $k$, if the value at the $k^{th}$ dimension of the mask $\m_k$ is 1 (\textit{resp.,} 0), we classify $\g_t^k$ as a dynamic Gaussian $\g_t^{k, {Dynamic}} \in \mathbb{R}^{D}$ (\textit{resp.,} a static Gaussian $\g_t^{k, {Static}} \in \mathbb{R}^{D}$).
For all static Gaussians $\{\g_t^{k,{Static}} \mid t = 2 : T\}$ starting from the second time step ($t = 2$), we directly replace them with the corresponding static Gaussians $\g_1^{k,{Static}}$ from the first time step ($t = 1$) within the same Gaussian index $k$. Hence, we obtain a static Gaussian set $\G_{static}=\{\g^{k,Static}|k=1:S\}$ and a dynamic Gaussian set $\cG_{Dynamic}=\{\g_t^{k,Dynamic}|_{t:1=T}^{k=1:K-S}\}$.
In this way, assuming we select $S$ static Gaussians, we can effectively reduce the total number of parameters that require updates at each time step from $K \times T$ to $S + (K - S) \times T$, thereby eliminating redundant updates for static components. Note that the number $S$ is learnable based on the mask. As a result, the number of tokens in our GVT is also learnable, making our method more versatile and adaptive to different content. The overall procedure is denoted as:

\begin{equation}
\begin{split}
\G_{Static}, \cG_{Dynamic} = \text{GSP}(\mathbb{\cG}), \cG \leftarrow \G_{Static} \cup \cG_{Dynamic}. \\
s.t.,
\begin{cases}
\G_{Static} =\{\g^{k,Static}|k=1:S\}, \\
\cG_{Dynamic} =\{\g_t^{k,Dynamic}|_{t:1=T}^{k=1:K-S}\}.
\end{cases}
\end{split}
\label{eq:gsp}
\end{equation}

We further introduce a constraint loss $\cL_{GSP}$ to regularize the proportion of dynamic Gaussians, encouraging the model to select fewer dynamic Gaussians (first term) and penalizing cases where the proportion exceeds the threshold $\tau$ (second term), controlled by the hyperparameters $\lambda_1$ and $\lambda_2$.

\begin{equation}
\cL_{GSP} = \lambda_1 \frac{1}{K}\sum_{k=1}^K \m_k 
+ \lambda_2 \operatorname{ReLU}\left( \frac{1}{K}\sum_{k=1}^K \m_k - \tau \right).
\label{eq:gsp_reg}
\end{equation}

\subsubsection{Vector Quantization for Coefficient Vector.}
\label{gvt_vq}
We quantize the 2D Gaussian set $\cG$ to map the continuous rendering parameters into discrete values. In this work, we omit additional parameters such as opacities and retain only the feature coefficients $\varsigma$ for each 2DGS (see \textit{Sec.~\ref{def_2dgs}}).  
We apply a VQ strategy~\cite{esser2021taming} using a codebook of size ${L} \times (D - 5)$ containing $L$ discrete codewords $\varsigma_\ell \in \mathbb{R}^{D - 5}$ (see Fig.~\ref{fig:framework} (e)). The feature coefficient $\varsigma_k$ of each Gaussian $\g_k$ is then quantized to its nearest codeword by: $\bar{\varsigma_k} = \varsigma_{\ell^*}, 
\ell^* = \arg\min_{\ell \in [1, {L}]} \, \| \varsigma - \varsigma_\ell \|$.
This results in the quantized 2D Gaussian set:  $\bar{\cG} = \{ \bar{\g}_k \mid k = 1 : S + (K - S) \times T \}$, where each quantized Gaussian $\bar{\g}_k \in \mathbb{R}^D$ contains the quantized coefficient vector $\bar{\varsigma}_k$. The procedure is denoted as:

\begin{equation}
\bar{\cG} = Q(\cG),  
s.t., 
\begin{cases}
\bar{\cG} = \{ \bar{\g}_k \mid k = 1 : S + (K - S) \times T \}, \\
\bar{\varsigma}_k = \varsigma_{\ell^*}, \varsigma_k \in \g_k, \bar{\varsigma_k} \in \bar{\g_k}, \\
\ell^* = \arg\min_{\ell \in [1, {L}]}  
\| \varsigma_k - \varsigma_\ell \|.
\end{cases}
\end{equation}

During optimization of VQ procedure, we employ the standard commitment loss 
$\mathcal{L}_{VQ}$ following VQGAN~\cite{esser2021taming} as
\begin{equation}
    \mathcal{L}_{VQ}
    =  \frac{1}{|\cG|}
    \sum_{k=1}^{|\cG|}
    \left\|
        \varsigma_k - \bar{\varsigma}_k
    \right\|^2_2, |\cG| = S + (K - S)\times T.
    \label{eq:commit}
\end{equation}

\subsection{Reconstruction and Optimization}
\subsubsection{Reconstruction} To render each individual token $\bar{\z}_t^i$ at spatial position $\mu_t^i \in \mathbb{R}^2$ and time step $t$, we follow~\cite{dong2025gaussiantoken} and aggregate the contributions of all quantized Gaussians $\bar{\g}_t^k$ in $\G_t$ based on their radiance. During the rendering, we duplicate all quantized static Gaussian primitives $\bar{\g}^{k,Static}$ for all time-step $t$ along with other quantized dynamic Gaussian primitives, $\bar{\g}^{k,Dynamic}_t$. Hence, we still have $\bar{\cG} =\{\bar{g}_t^k |_{t=1:T}^{i=1:K}\}$ for performing rasteraization.
To this end, the radiance weight $\pi_t^k$ from a Gaussian $\g_t^k$, located at position $\mu_t^k$ to position $\mu_t^i$, is computed as:
$\pi_t^k = \exp\left(-\frac{1}{2}(\mu_t^i - \mu_t^k)^T \Sigma^{-1} (\mu_t^i - \mu_t^k)\right)$, which will be applied to the quantized coefficient vectors $\bar{\varsigma}_t^k$ (see \textit{Sec.~\ref{gvt_vq}}), resulting in:
$
\bar{\z}_t^i = \sum_{k=1}^K \pi_t^k \, \bar{\varsigma}_t^k.
$
The overall procedure to render the video token $\bar{\z}_t^i$ is defined as:

\begin{equation}
\bar{\z}_t^i = \sum_{k=1}^K \exp\left(-\frac{1}{2}(\mu_t^i - \mu_t^k)^T \Sigma^{-1} (\mu_t^i - \mu_t^k)\right) \bar{\varsigma}_t^k. 
\label{eq:render}
\end{equation}

After rasterization, we obtain the results for all video tokens  
$\mathbb{\bar{Z}} = \{ \bar{\z}_t^i |_{t = 1 : T}^{i = 1 : N} \}$.  
We then apply a decoder $\cD(\cdot)$ to reconstruct the video:  
$\bar{\mathbb{V}} = \cD(\mathbb{\bar{Z}})$. As mentioned in \textit{Sec.~\ref{def_vt}}, we minimize the reconstruction loss $\mathcal{L}_{Recon}$ by resolving:

\begin{equation}
\mathcal{L}_{Recon} = \left\| \mathbb{V} - \cD(\bar{\mathbb{Z}}) \right\|, s.t., \bar{\mathbb{Z}}=\{\bar{\z}_t^i |_{t=1:T}^{i=1:N}\}.
\label{eq:decode}
\end{equation}

To encourage perceptually realistic video reconstructions, we further adopt
a generator loss $\mathcal{L}_{Gen}$, following VQGAN~\cite{esser2021taming},
with a discriminator $\mathcal{T}(\cdot)$ defined as:
\begin{equation}
\mathcal{L}_{Gen}
= \mathbb{E}_{\mathbb{V}} \big[ \log \big( 1 - \mathcal{T}(\cD(\bar{\mathbb{Z}})) \big) \big],
\label{eq:adv}
\end{equation}
where $\mathcal{T}(\cdot)$ is optimized with 
$
\mathcal{L}_{Adv}
= - \mathbb{E}_{\mathbb{V}} \big[ \log \mathcal{T}(\mathbb{V}) \big]
  - \mathbb{E}_{\mathbb{V}} \big[ \log \big( 1 - \mathcal{T}(\cD(\bar{\mathbb{Z}})) \big) \big]
$, as an adversarial loss as in VQGAN~\cite{esser2021taming}.

\subsubsection{Objective Function.} 
Overall, our model is optimized using the reconstruction loss 
(\ie, Eq.~\ref{eq:decode}), the generator loss (\ie, Eq.~\ref{eq:adv}), 
the GSP regularization term (\ie, Eq.~\ref{eq:gsp_reg}), and the VQ loss (\ie, Eq.~\ref{eq:commit}). The overall objective can be presented as:
\begin{equation}
\cL = \cL_{Recon} + \alpha \cL_{Gen} + \cL_{GSP} + \beta \cL_{VQ}.
\label{eq:loss}
\end{equation}
\section{Experiments}
\subsection{Experimental Protocols}
\subsubsection{Datasets.} 

We use three video datasets:
On \textit{UCF101}, following~\cite{yulanguage}, we resize all videos to a resolution of ($128 \times 128$). The training set (9,537 videos) is used for model training and reconstruction and generation evaluation, consistent with~\cite{yulanguage}, whereas its test set (3,783 videos) is used to evaluate action recognition and compression.
On \textit{Kinetics}, following~\cite{yulanguage}, we resize all videos to a resolution of ($128 \times 128$). The training set of Kinetics-600 (\ie, K600) subset with 600 classes is used for model training and reconstruction evaluation, consistent with~\cite{yulanguage}, whereas its test set is used for compression evaluation. The validation set of Kinetics-400 (\ie, K400) subset with 400 classes is used to evaluate action recognition, consistent with~\cite{tong2022videomae}.
On \textit{DAVIS},
following~\cite{sun2024splatter}, we adopt the DAVIS-2017 subset at 480P resolution of $854 \times 480$ with 60 training videos and 30 validation videos to train and evaluate video representation tasks in comparison with NeRF- and GS-based methods.

\subsubsection{Implementation Details.}
We build our GVT on top of the baseline MAGVIT2~\cite{yulanguage}\footnote{As the original source of MAGVIT-v2~\cite{yulanguage} is unavailable, we adopt the widely recognized public reproduction of Open-MAGVIT-v2~\cite{luo2024open}.}. We first extract the latent video representations $\mathbb{Z}$ from a $128 \times 128 \times 17$ video clip and then apply our GVT to project and quantize into Gaussian Video Tokens $\bar{\cG}$.
We set the dimensionalities of the initial Query, Gaussian Set, and Mask Query to $D_1 = 64$, $D_2 = 69$, and $D_3 = 64$. The number of Gaussians is initialized as $K \times T = 512 \times 5$ for each $128 \times 128 \times 17$ video clip. After processing through STGE, which consists of $B = 3$ DSTF blocks, followed by the GSP, we obtain an average of $1,868$, $1,964$, and $60,132$ Gaussian tokens for inputs of size $128 \times 128 \times 17$, $128 \times 128 \times 17$, and $854 \times 480 \times 17$ across all videos in the UCF101, K600, and DAVIS datasets. Each final token has a size of $D = 13$ with 8-dimensional coefficients (see \textit{Sec.~\ref{def_2dgs}}), so we adopt a $4096 \times 8$ codebook.
\begin{table}[!t]
\centering
\caption{Reconstruction results in rFVD on UCF101 and K600. $^*$ denotes that our token number is learnable. Blue indicates \textbf{Top-3} methods: darkest ($1^{st}$), medium ($2^{nd}$), light ($3^
{rd}$).}
\begin{tabular}{l|cc|cc}
\noalign{\hrule height 1pt}
\textbf{Methods}         & \textbf{\#Tokens} & \begin{tabular}[c]{@{}l@{}}\textbf{Token} \\\textbf{Size}\end{tabular} & \begin{tabular}[c]{@{}l@{}}\textbf{UCF101}\\ \textbf{(rFVD$\downarrow$)}\end{tabular} & \begin{tabular}[c]{@{}l@{}}\textbf{K600}\\ \textbf{(rFVD$\downarrow$)}\end{tabular} \\ 
\noalign{\hrule height 1pt}
MaskGIT~\cite{chang2022maskgit}        & 4352       & 256            & 240                                                      & 202                                                   \\
VQGAN~\cite{esser2021taming}          & 4352       & 256              & 299                                                      & 270                                                   \\
TATS~\cite{ge2022long}           & 1024       & 256            & 162                                                      & -                                     \\
MAGVIT~\cite{yu2023magvit}         & 1024       & 256            & 25                                                       & -                                      \\
OmniTok~\cite{wang2024omnitokenizer}        & 1280       & 8               & 107                                                     & 84                                                    \\
LARP-L~\cite{wanglarp}         & 1024       & 8            & \topthree{20}               & \toptwo{13}    \\
SweetTok~\cite{tan2024sweettok}         & 1280       & 256            & \topthree{20}               & 25    \\
MAGVIT-v2~\cite{yulanguage}        & 1280       & 18             & \toptwo{16.7}                                                     & \topthree{24.3}                                                  \\ 
\noalign{\hrule height 1pt}
GVT (UCF101) & 1868$^*$       & 13        & \topone{\textbf{12.6}}  & -                                                    \\
GVT (K600)    & 1964$^*$      & 13          & -     & \topone{\textbf{8.6}}        \\ \noalign{\hrule height 1pt}                                      
\end{tabular}
\label{tab:VR-1}
\end{table}

We train three separate models to perform the reconstruction task on the UCF101, K600, and DAVIS-2017-480P datasets. Following~\cite{luo2024open}, we first pre-train our model on ImageNet-1K~\cite{imagenet} at a resolution of $128 \times 128$ for 50 epochs using a base learning rate of $1 \times 10^{-4}$. We then fine-tune this model on UCF101 and K600 at the same resolution for 50 and 10 epochs, respectively, both with a base learning rate of $1 \times 10^{-4}$, resulting in two domain-specific models.
For the DAVIS dataset, we further fine-tune the model trained on K600 for an additional 20 epochs at a higher resolution of $854 \times 480$, again using a base learning rate of $1 \times 10^{-4}$. 
%
All training procedures are conducted using 8 NVIDIA A100 80GB GPUs and optimized with the Adam optimizer and the unified hyper-parameters are $\alpha = 0.1$, $\beta = 0.25$, $\lambda_1 = 5 \times 10^{-3}$, $\lambda_2 = 2 \times 10^{-2}$, and $\tau = 0.25$.

\subsection{Experimental Results}
\subsubsection{Video Reconstruction.} We evaluate the domain-level video reconstruction capability on two benchmarks, UCF101 and K600. We compare against several video tokenizers, including MaskGIT~\cite{chang2022maskgit}, VQGAN~\cite{esser2021taming}, TATS~\cite{ge2022long}, MAGVIT~\cite{yu2023magvit}, OmniTok~\cite{wang2024omnitokenizer}, LARP-L~\cite{wanglarp}, SweetTok~\cite{tan2024sweettok}, and MAGVIT-v2~\cite{yulanguage}. We use rFVD~\cite{unterthiner2018accurate} to evaluate domain-level reconstruction quality.
Table~\ref{tab:VR-1} shows that GVT achieves the best reconstruction performance, while maintaining a reasonable token count (\ie, \#Tokens) and token size. It is worth noting that, among all methods, our GVT is the only method that supports a learnable token count, enabling greater adaptability. We also provide the qualitative reconstruction results in Fig.~\ref{fig:vis-rec}.

As shown in Table~\ref{tab:VR-2}, we further evaluate the video representation task on the DAVIS-2017-480P benchmark and compare our GVT against Radiance Field-based approaches, including NERV~\cite{chen2021nerv}, HERV~\cite{chen2023hnerv}, 4DGS~\cite{Wu_2024_CVPR}, RoDynRF~\cite{liu2023robust}, Deformable Sprites~\cite{ye2022deformable}, OmniMotion~\cite{wang2023tracking}, CoDeF~\cite{ouyang2024codef}, and Splatter-a-video~\cite{sun2024splatter}.
Unlike domain-level reconstruction tasks, the video representation task emphasizes instance-level fidelity. Therefore, we adopt SSIM and LPIPS as perceptual metrics.
GVT is the only Radiance Field–based approach that operates in a feed-forward manner, requiring no additional per-video fitting.
Despite this, it still achieves the best LPIPS and the third-best SSIM among all competitors.

\subsubsection{Video Action Recognition.}
To evaluate whether our representation method preserves essential semantic information, we conduct video action recognition experiments.
Specifically, we use the GVT and MAGVIT-v2 models trained based on the UCF101 and K600 training sets to reconstruct videos from the UCF101 test set and the K400 validation set.
The reconstructed videos are then fed into the state-of-the-art video action recognition method, VideoMAE~\cite{tong2022videomae}.
We also include results obtained using the raw (\ie, uncompressed) videos as the baseline.
As shown in Table~\ref{tab:VAR}, the reconstructed videos from GVT achieve higher Top-1 and Top-5 accuracy on both UCF101 and K400 compared to MAGVIT-v2, demonstrating that our method better preserves semantic information.

\subsubsection{Video Compression.} 


\begin{table}[!t]
\centering
\caption{Representation results on DAVIS-2017-480P. Our method is the only feed-forward method. Blue indicates \textbf{Top-3} methods: darkest ($1^{st}$), medium ($2^{nd}$), light ($3^
{rd}$).}

\begin{tabular}{l|c|cc}
\noalign{\hrule height 1pt}
\textbf{Methods} & \textbf{Fitting Time} & \textbf{SSIM$\uparrow$} & \textbf{LPIPS$\downarrow$} \\
\noalign{\hrule height 1pt}
NeRV~\cite{chen2021nerv}                  & $\sim$45 mins        & 0.72                         & 0.312 \\
HNeRV~\cite{chen2023hnerv}                & $\sim$15 mins        & 0.72                         & {0.252} \\
4DGS~\cite{Wu_2024_CVPR}                  & $\sim$40 mins        & 0.57                         & 0.394 \\
RoDynRF~\cite{liu2023robust}              & \textgreater 24 hrs  & 0.72                         & 0.394 \\
Deformable Sprites~\cite{ye2022deformable}& $\sim$30 mins        & 0.70                         & 0.301 \\
OmniMotion~\cite{wang2023tracking}        & \textgreater 24 hrs  & 0.72                         & 0.371 \\
CoDeF~\cite{ouyang2024codef}              & $\sim$30 mins        & \toptwo{0.82}                & 0.29 \\
Splatter-a-Video~\cite{sun2024splatter}   & $\sim$30 mins        & \topone{\textbf{0.84}}       & \topthree{0.228} \\
MAGVIT-v2~\cite{yulanguage}                                       & {Feed-Forward} & {0.74}             & \toptwo{\textbf{0.140}} \\
\noalign{\hrule height 1pt}
GVT                                       & {Feed-Forward} & \topthree{0.78}             & \topone{\textbf{0.129}} \\
\noalign{\hrule height 1pt}
\end{tabular}
\label{tab:VR-2}
\end{table}

To further validate the compactness of our GVT, we conduct video compression experiments on the UCF101 and K600 test sets. For variable-rate compression, different quantization step sizes are applied to the reconstructed video tokens $\bar{\mathbb{Z}}$ (see Eq.~\ref{eq:render} and Eq.~\ref{eq:decode} for the generation of $\bar{\mathbb{Z}}$). The quantized video tokens are then losslessly compressed using a hybrid neural entropy model with context and hyperprior architectures following~\cite{cheng2020learned}. To improve lossless compression efficiency, we fine-tune the auxiliary entropy model based on the UCF101 and K600 training sets using a rate--distortion (RD) objective, with all modules in the GVT framework kept frozen throughout optimization.

\begin{figure*}[t!]
  \centering
 \caption{Rate--Distortion performance comparison between our and other methodologies on UCF101 and K600.}
  \begin{minipage}{0.24\textwidth}
    \centering
    \includegraphics[width=\linewidth]{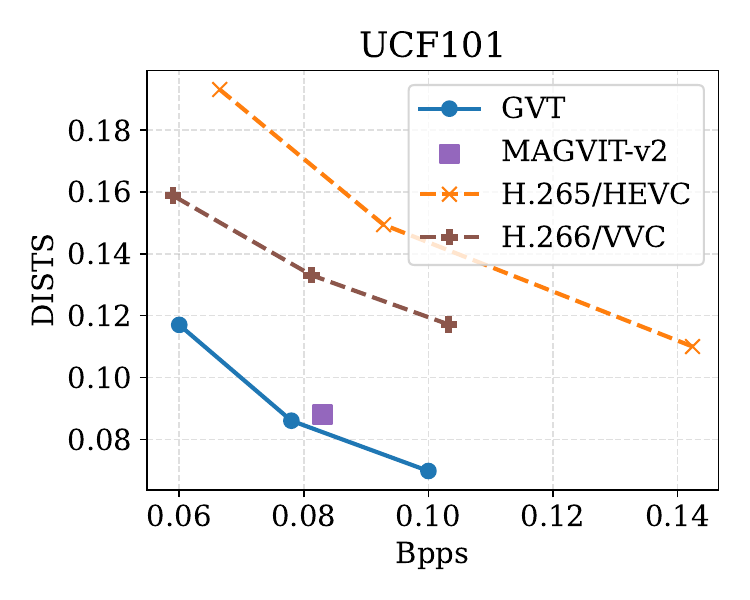}
  \end{minipage}
  \hfill
  \begin{minipage}{0.24\textwidth}
    \centering
    \includegraphics[width=\linewidth]{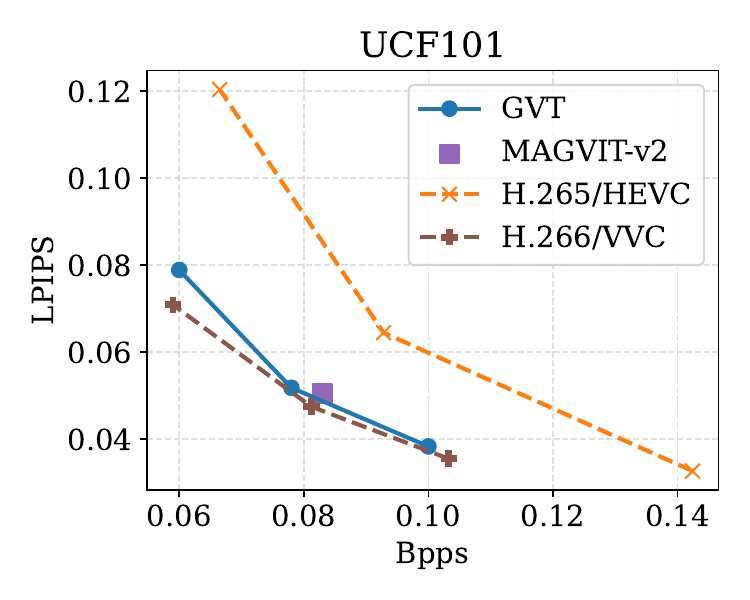}
  \end{minipage}
  \hfill
  \begin{minipage}{0.24\textwidth}
    \centering
    \includegraphics[width=\linewidth]{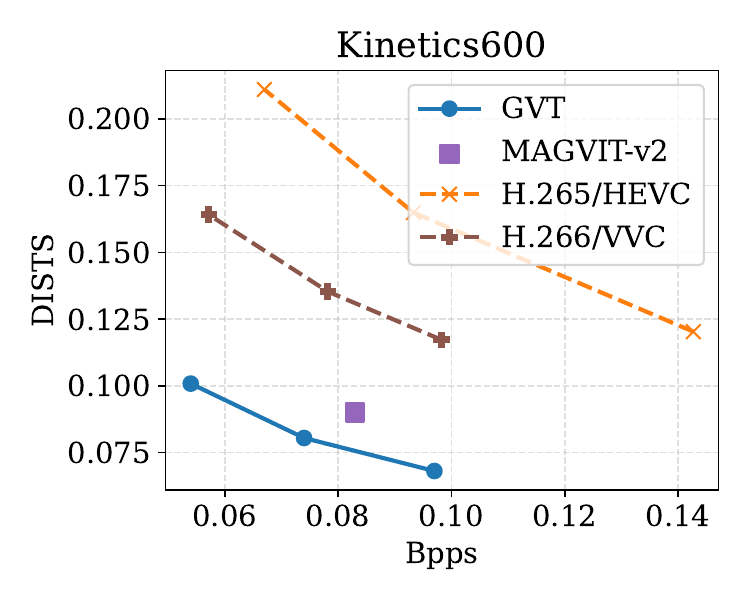}
  \end{minipage}
  \hfill
  \begin{minipage}{0.24\textwidth}
    \centering
    \includegraphics[width=\linewidth]{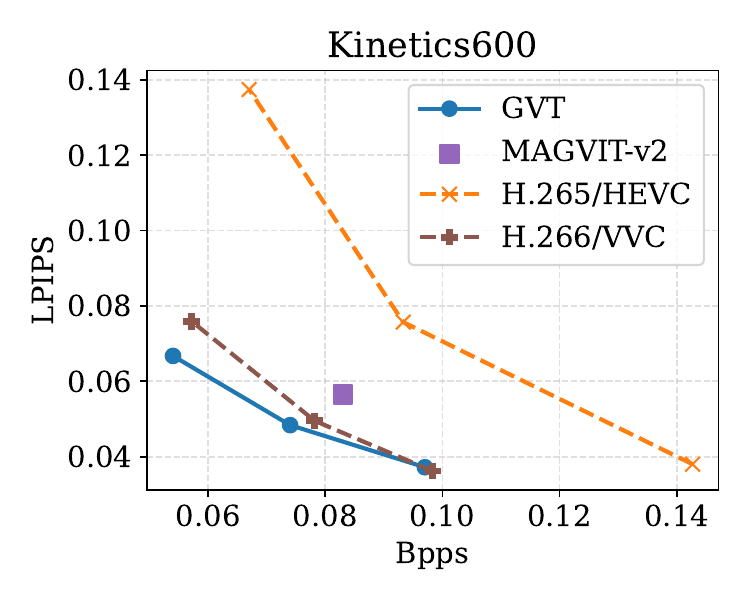}
  \end{minipage}
  \label{fig:vc}
\end{figure*}

\begin{figure*}
  \centering
    \caption{Reconstruction results on UCF101 with corresponding error maps in $2^{nd}$ row (brighter regions indicate larger errors).}
    \includegraphics[width=\linewidth]{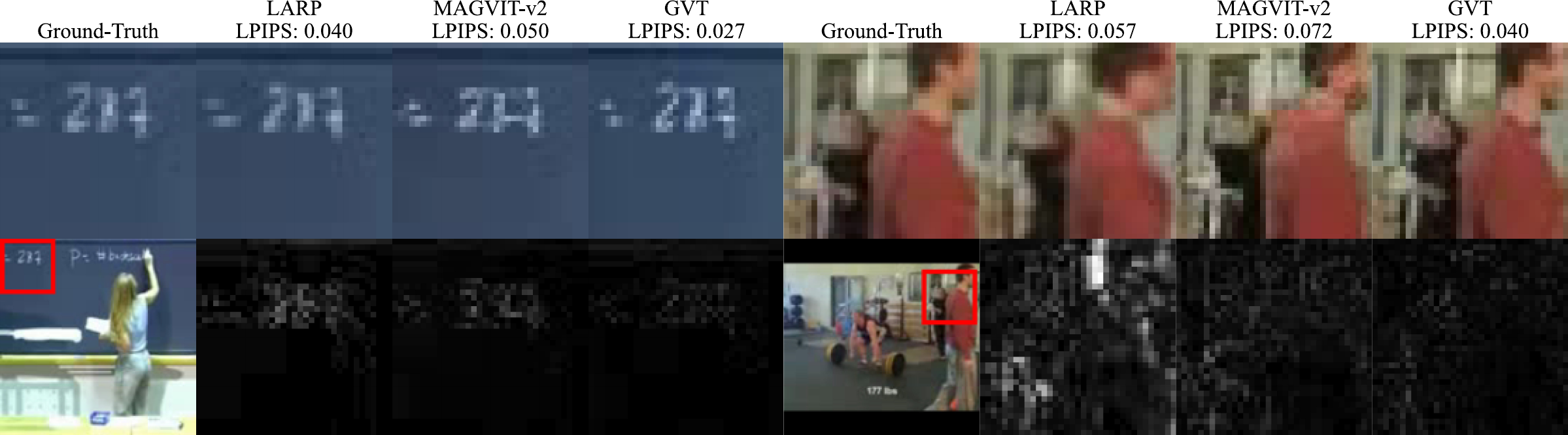}
    \label{fig:vis-rec}
\end{figure*}

We report the RD curves in Fig.~\ref{fig:vc}, where conventional video codecs H.266/VVC~\cite{bross2021overview} and H.265/HEVC~\cite{sullivan2012overview}, implemented using the reference software {VTM-23.4~\cite{vtm} and HM-18.0~\cite{HM}}, respectively, are used as anchors. Following recent perceptual video compression studies~\cite{qi2025generative, zhenggenerative}, we adopt DISTS and LPIPS as the perceptual reconstruction metrics. The results show that our method achieves superior RD performance, yielding average bit-rate reductions of {54.8}\% and {2.1}\% in terms of DISTS and LPIPS, respectively, compared with the latest video compression standard H.266/VVC.


\subsubsection{Video Generation.}
We further investigate the unconditional video generation capability of GVT. Due to the absence of large-scale pre-trained cross-modal alignment models for video-based 2D Gaussian Splatting, together with the continuous and unordered nature of 2D Gaussian primitives, we restrict our study to unconditional video generation, following a denoising-based generative setting~\cite{he2022lvdm}. Specifically, we adopt the recent flow-matching paradigm of~\cite{li2025back} in the rigid latent space $\mathbb{Z}$ (see Eq.~\ref{eq:encode}) before projecting it into 2D Gaussian primitives $\mathcal{G}$.
Given a clean latent $\mathbb{Z}$, we sample Gaussian noise $\epsilon \sim \mathcal{N}(0,\mathbf{I})$ to construct a noisy latent $\mathbb{Z}_t = t\mathbb{Z} + (1-t)\epsilon$ at time step $t \in [0,1]$. The corresponding target velocity field is defined as $\mathbb{F}_t = (\mathbb{Z}-\mathbb{Z}_t)/(1-t)$. We employ an auxiliary prediction network to estimate the clean latent $\hat{\mathbb{Z}}_t$ from $\mathbb{Z}_t$, and compute the predicted velocity as $\hat{\mathbb{F}}_t = (\hat{\mathbb{Z}}_t-\mathbb{Z}_t)/(1-t)$. We optimize this network with an $\ell_2$ loss on the velocity field, while all modules in the GVT framework remain frozen.
At inference time, generation starts from Gaussian noise and progressively integrates the latent trajectory from $t=1$ to $t=0$ using an Euler ordinary differential equation (ODE) solver. At each step, the prediction network estimates the clean latent and its corresponding velocity field, which is then used to update the current latent state. The final latent $\mathbb{Z}$ is subsequently fed into GVT to synthesize the video.

\begin{figure*}
  \centering
    \caption{Qualitative results of unconditional video generation produced by our GVT model trained on UCF101.}\includegraphics[width=\linewidth]{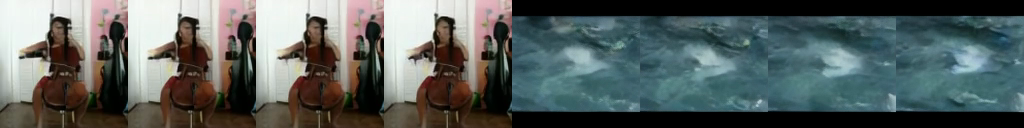}
    \label{fig:vis-gen}
\end{figure*}

We report unconditional video generation performance on UCF101 in Table~\ref{tab:fvd_compare}, where GVT is compared with several representative video generation approaches. For a fair comparison, we further equip MAGVIT-v2 with the same flow-matching paradigm in the rigid latent space $\mathbb{Z}$, denoted as MAGVIT-v2$^*$. The results show that GVT outperforms the MAGVIT-v2$^*$ baseline and achieves performance comparable to several existing methods, including MCVD~\cite{voleti2022mcvd} and TGAN-v2~\cite{saito2020train}. We also present qualitative results in Fig.~\ref{fig:vis-gen}, which show two generated videos, a person playing the cello and waves. These results demonstrate the generation capability of GVT, although a performance gap remains compared with state-of-the-art generation methods.



\begin{table}[t!]
\centering
\caption{Action recognition results using VideoMAE on raw videos and videos reconstructed by MAGVIT-v2 and GVT.}
\begin{tabular}{l|cccc}
\noalign{\hrule height 1pt}
            & \multicolumn{2}{c}{\textbf{UCF101}}      & \multicolumn{2}{c}{\textbf{K400}}       \\
\textbf{Methods}           & \textbf{Top-1$\uparrow$}   & \textbf{Top-5$\uparrow$}  & \textbf{Top-1$\uparrow$}  & \textbf{Top-1$\uparrow$}  \\
\noalign{\hrule height 1pt}
Raw Video   & 91.25  & 98.55 & 81.47 & 95.04 \\
MAGVIT-v2~\cite{yulanguage} & 85.73  & 97.15 & 76.88 & 92.64 \\
\noalign{\hrule height 1pt}
GVT        & 86.60  & 97.49 & 78.05 & 93.26 \\ 
\noalign{\hrule height 1pt}
\end{tabular}
\label{tab:VAR}
\end{table}

\begin{table}[!t]
\centering
\caption{Unconditional video generation results on UCF101.} 
\begin{tabular}{l|c}
\noalign{\hrule height 1pt}
\textbf{Methods} & \textbf{FVD$\downarrow$}  \\
\noalign{\hrule height 1pt}
MCVD~\cite{voleti2022mcvd}       & 1143 \\
TGAN-v2~\cite{saito2020train}     & 1209 \\
TATS~\cite{ge2022long}        & 420  \\
MoCoGAN~\cite{tian2021good}  & 700  \\
LVDM~\cite{he2022lvdm}       & 372  \\
MAGVIT-v2$^*$~\cite{yulanguage}      & 3772 \\ \noalign{\hrule height 1pt}
GVT        & 1240 \\
\noalign{\hrule height 1pt}
\end{tabular}
\label{tab:fvd_compare}
\end{table}

\begin{table}[!t]
\centering
\caption{Ablation study of each module on UCF101.} 
\begin{tabular}{l|c|c}
\noalign{\hrule height 1pt}
\textbf{Methods}     & \textbf{\#Tokens} & \textbf{rFVD$\downarrow$}  \\
\noalign{\hrule height 1pt}
GVT w/o (GSP \& STA)      & 2560      & 18.3 \\
GVT w/o GSP & 2560      & 9.2  \\
GVT w fixed partitioning   & 1868      & 40   \\ \noalign{\hrule height 1pt}
GVT  & 1868      & 12.6 \\
\noalign{\hrule height 1pt}
\end{tabular}
\label{tab:ab}
\end{table}


\begin{figure}
  \centering
    \caption{Visualization for the distribution of the \textcolor{DeepSkyBlue2}{static} and the \textcolor{red}{dynamic} Gaussians (\ie, the $2^{nd}$ column), along with the radiance field maps showing rasterization weights at each spatial position (\ie, the $3^{rd}$ column).}
    \includegraphics[width=\linewidth]{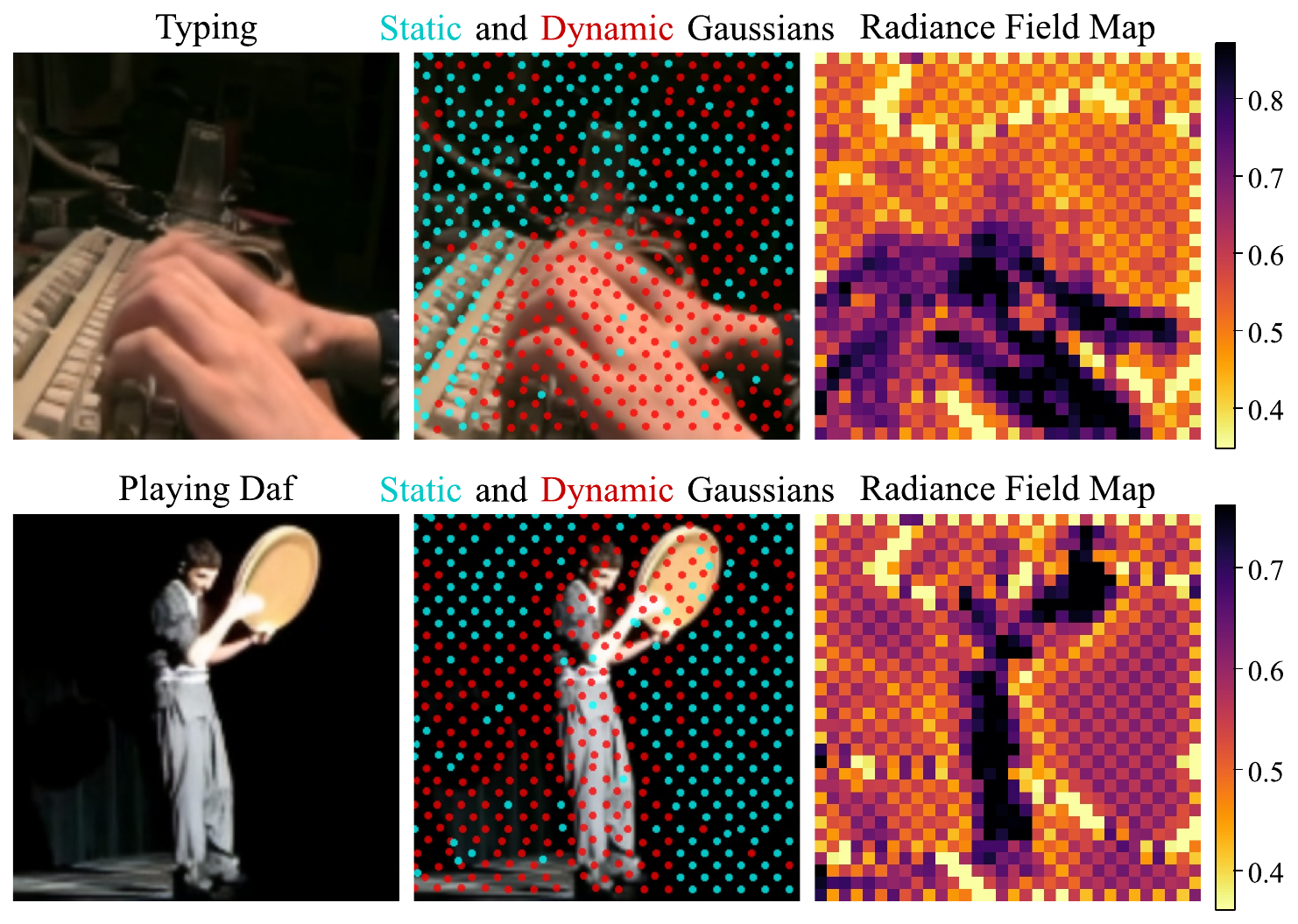}
    \label{fig:vis-dis}
\end{figure}

\subsection{Quantitative and Qualitative Analysis}
\subsubsection{Ablation Study.} We conduct an ablation study on the UCF101 dataset to evaluate the effectiveness of each module in GVT in Table.~\ref{tab:ab}.
We begin with a baseline variant, GVT w/o (GSP \& STA), which replaces all STA modules (Fig.~\ref{fig:framework} (b)) with a spatial-only attention module, which exploits only spatial correlations, and does not use GSP to partition dynamic and static Gaussians.
Adding STA (the $2^{nd}$ row) reduces rFVD by 49.7\%, highlighting the benefit of modeling temporal correlations.
Similarly, further incorporating GSP (the $4^{th}$ row) reduces rFVD by 31.1\% while saving 27.0\% tokens from the baseline, demonstrating its ability to reduce temporal redundancy.
These results clearly show the importance of explicitly modeling temporal correlations and eliminating redundancy to ensure both the effectiveness and efficiency of our model. Moreover, to further validate the effectiveness of GSP, we evaluate a variant with a fixed partitioning strategy (the $3^{rd}$ row). In this variant, we directly select a subset (matching the token count of our complete method) from the initial Gaussians $\cG_{\text{init}}$ as static Gaussians prior to optimization.
This leads to a substantial performance drop, with rFVD increasing by $2.2\times$ compared to our GVT.

\subsubsection{Analysis of Spatial and Temporal Versatility.} We provide two visualizations in Fig.~\ref{fig:vis-dis} to further illustrate and analyze the spatial and temporal versatility of our approach, using videos of the ``playing Daf" and ``typing" actions.
Specifically, we visualize the distribution of static and dynamic Gaussians by respectively displaying their locations $\mu_k$ in red and blue in the $2^{nd}$ column. For static background regions, most Gaussians are classified as static, whereas for moving objects, such as the typing fingers and the human body, the majority of Gaussians are classified as dynamic. Moreover, in the $3^{rd}$ column, we visualize the Radiance Field Map, which displays the rasterization weights aggregated from the 2D Gaussians at each spatial position (see Eq.~\ref{eq:render}).
The visualization reveals that objects containing richer information (\eg, the human body and hands) receive higher rasterization weights from 2D Gaussian primitives.
Such rasterization characteristics enable the reconstructed latent $\bar{\mathbb{Z}}$ to preserve both spatial and temporal versatility with greater compactness, thereby improving storage efficiency for video compression and reconstruction quality for downstream tasks such as video synthesis and video action recognition.


\section{Conclusion}
In this work, we introduced GVT, a 2DGS-based video representation method that exploits 2DGS to learn compact and semantically rich video representations. By representing videos with 2DGS generated in a feed-forward manner, GVT preserves both spatial and temporal versatility, resulting in state-of-the-art reconstruction and compression performance, as well as competitive action recognition and generation performance. Overall, this work highlights the effectiveness of Gaussian Splatting-based latent representations as a promising direction for efficient video representation and facilitates future research in video representation and Gaussian Splatting.

\textbf{Limitations and Future Works.} This study is currently restricted to unconditional video generation, and its performance still lags behind state-of-the-art methods due to the absence of pre-trained multimodal models and the inherent properties of 2DGS. Our generation result should be viewed as preliminary evidence that GVT tokens can support generative modeling. In the future work, we plan to develop effective alignment methods for video-based 2DGS pipelines to support more advanced conditional video generation tasks, such as text-to-video, thereby broadening their applicability to next-generation video foundation models.

\begin{acks}
This work is supported in part by the Research Grants Council (RGC) of the Hong Kong SAR under the General Research Fund (17203023), the Collaborative Research Fund (C5052-23G), and the NSFC/RGC Collaborative Research Scheme (CRS\_HKU703/24). The research work described in this paper was conducted in the JC STEM Lab of Multimedia and Machine Learning funded by The Hong Kong Jockey Club Charities Trust.
\end{acks}

\bibliographystyle{ACM-Reference-Format}
\bibliography{sample-base}
\appendix


\section{Appendix}
\subsection{Methodological Details}
\label{sec:method}
\subsubsection{Spatio-Temporal Attention}
\label{sec:gvt_sta}
The STA module (Fig. 2 (b)) is extended from the standard visual self-attention mechanism~\cite{dosovitskiy2021vit}.  
Specifically, we take the joint feature 
$\cI \in \mathbb{R}^{T \times N \times F}$ as input, where $T$, $N$, and $F$ denote the temporal, spatial, and channel dimensions, respectively. We first split $\cI$ along the temporal dimension into $T$ feature matrices 
$\cI = \{ \I_t \mid t = 1 : T \}.$
For each temporal slice $\I_t$, we directly apply self-attention~\cite{dosovitskiy2021vit} to enhance the features in the spatial dimension.  
Next, we aggregate the updated temporal slices back into the joint feature $\cI$ and split it along the spatial dimension: $\cI = \{ \I_i \mid i = 1 : N \}$. 
For each spatial slice $\I_i$, we again apply self-attention to enhance the features in the temporal dimension.  
Finally, we aggregate all updated spatial slices $\I_i$ back into the joint feature $\cI$, obtaining the refined spatio-temporal representation.

\subsubsection{Deformable Spatio-Temporal Fusion}
\label{sec:gvt_dstf}
We introduce the DSTF module (Fig.~2 (c)), inspired by the 2D Gaussian embedding strategy proposed in \cite{dong2025gaussiantoken}. This module uses the query features to project information from the latent representation into the most relevant Gaussian and then updates the query accordingly. Specifically, it takes three inputs: the Gaussian tensor $\cG \in \mathbb{R}^{T \times K \times D_2}$, the query tensor $\cQ \in \mathbb{R}^{T \times K \times D_1}$, and the latent tensor $\cZ \in \mathbb{R}^{T \times K \times F}$.
We first apply MLPs along the channel dimension of $\cG$ to align it with the channel dimension of $\cQ$, enabling element-wise addition and subsequent STA processing to produce an updated query tensor $\cQ$. Meanwhile, we extract the Gaussian positions (consisting of two coordinates $x, y$) from $\cG$ as the reference tensor $\mu \in \mathbb{R}^{T \times K \times 2}$.

Next, we split these three tensors $\mu$, $\cQ$, and $\cZ$ along the temporal dimension into $T$ slices: $\mu_t \in \mathbb{R}^{K \times 2}$, $\Q_t \in \mathbb{R}^{K \times D_1}$, and $\Z_t \in \mathbb{R}^{K \times F}$, where $\mu_t$, $\Q_t$, and $\Z_t$ are used as the reference, query, and key-value matrices, respectively. We then apply the deformable cross-attention mechanism~\cite{zhu2021deformable} at each time step $t$ to compute the updated query $\Q_t \in \mathbb{R}^{K \times D_1}$.
The updated $\Q_t$ tensors are aggregated back into the full query tensor $\cQ$, which is further refined by STA modules to obtain the final query representation $\cQ$. Meanwhile, the Gaussian tensor $\cG = \{\g_t^k\ |_{t=1:T}^{k=1:K}\}$ is also updated via element-wise addition with the residual features $\Delta \cG = \{\Delta \g_t^k\ |_{t=1:T}^{k=1:K}\}$, which are obtained by feeding the produced final query representation $\cQ$ through skip connections and standard MLP layers. The final outputs are the updated Gaussian tensor $\cZ$ and query tensor $\cQ$.

\subsubsection{Temporal Alignment.}
\label{sec:gvt_order}
Most conventional video tokenizers maintain a fixed spatial index alignment across time steps. Each frame is divided into a uniform grid of patches, indexed in row-major order. The patch at position $\mu_t^i$ corresponding to token $\bar{\z}_t^i$ at time-step $t$ always shares the same spatial index $i$ as the patch at $\mu_{t-1}^i$ at time-step $t-1$, ensuring temporal alignment of spatial indices throughout the video.
In contrast, our GVT represents videos using 2D Gaussian
primitives, where positions are continuous and naturally unordered. 

To address this, we align their spatial indices across time steps during initialization.
Specifically, during the initialization of $\cG_{init}$ (see Fig.~2 (a) and \textit{Sec.~3.2.2}), we first initialize a Gaussian matrix $\G_{init} \in \mathbb{R}^{K \times D_2}$ with $K$ 2D Gaussian vectors $\g_k \in \mathbb{R}^{D_2}$ (\ie, $\G_{init} = \{\g_k \mid k = 1:K\}$). We then expand $\G_{init}$ by duplicating it $T$ times along the temporal dimension to obtain $\cG_{init} \in \mathbb{R}^{T \times K \times D_2}$. Therefore, at the beginning, for each time step $t$, we have $\g_{t-1}^k = \g_t^k$, ensuring that the Gaussians are initially ``aligned" across time-steps. 
Similarly, we align the initial query tensor $\cG_{Query} \in \mathbb{R}^{T \times K \times D_1}$ and mask query tensor $\cG_{Mask} \in \mathbb{R}^{T \times K \times D_3}$ by first initializing $\G_{Query} \in \mathbb{R}^{K \times D_1}$ and $\G_{Mask} \in \mathbb{R}^{K \times D_3}$, and then duplicating them along the temporal dimension.

\begin{figure}[t!]
  \centering
    \caption{Visualization of our Temporal Alignment Strategy and the principle behind GSP. In \textbf{the $1^{st}$ row}, we initialize the 2D Gaussians $\cG \in \mathbb{R}^{T \times K \times D_2}$ by duplicating the same initial Gaussian set $\G \in \mathbb{R}^{K \times D_2}$ across all $T$ time steps, resulting in identical Gaussian positions at $t{=}1$ and $t{=}2$.
    The {$2^{nd}$ row} visualizes the residuals (\ie, the update patterns) produced during the Spatio-Temporal Gaussian Embedding (STGE) stage, indicating how the Gaussians are updated through STGE (see Fig.~2(a)).
    Our GSP module is designed to identify Gaussians according to the similarity of their update patterns, as illustrated in the $2^{nd}$ row. Gaussians that exhibit similar residuals across time-steps $t{=1}$ and $t{=}2$ are highlighted with \textcolor{DeepSkyBlue2}{blue} arrows and classified as \textcolor{DeepSkyBlue2}{Static} Gaussians, as they result in minimal temporal differences in their updated positions. (\ie, the $3^{rd}$ row). In contrast, Gaussians with diverse or significant update patterns are marked with \textcolor{red}{red} arrows and classified as \textcolor{red}{Dynamic} Gaussians, showing greater displacement across time in the updated Gaussians. For improved clarity, only the most distinguishable points are shown in the visualization.}
  \includegraphics[width=\linewidth]{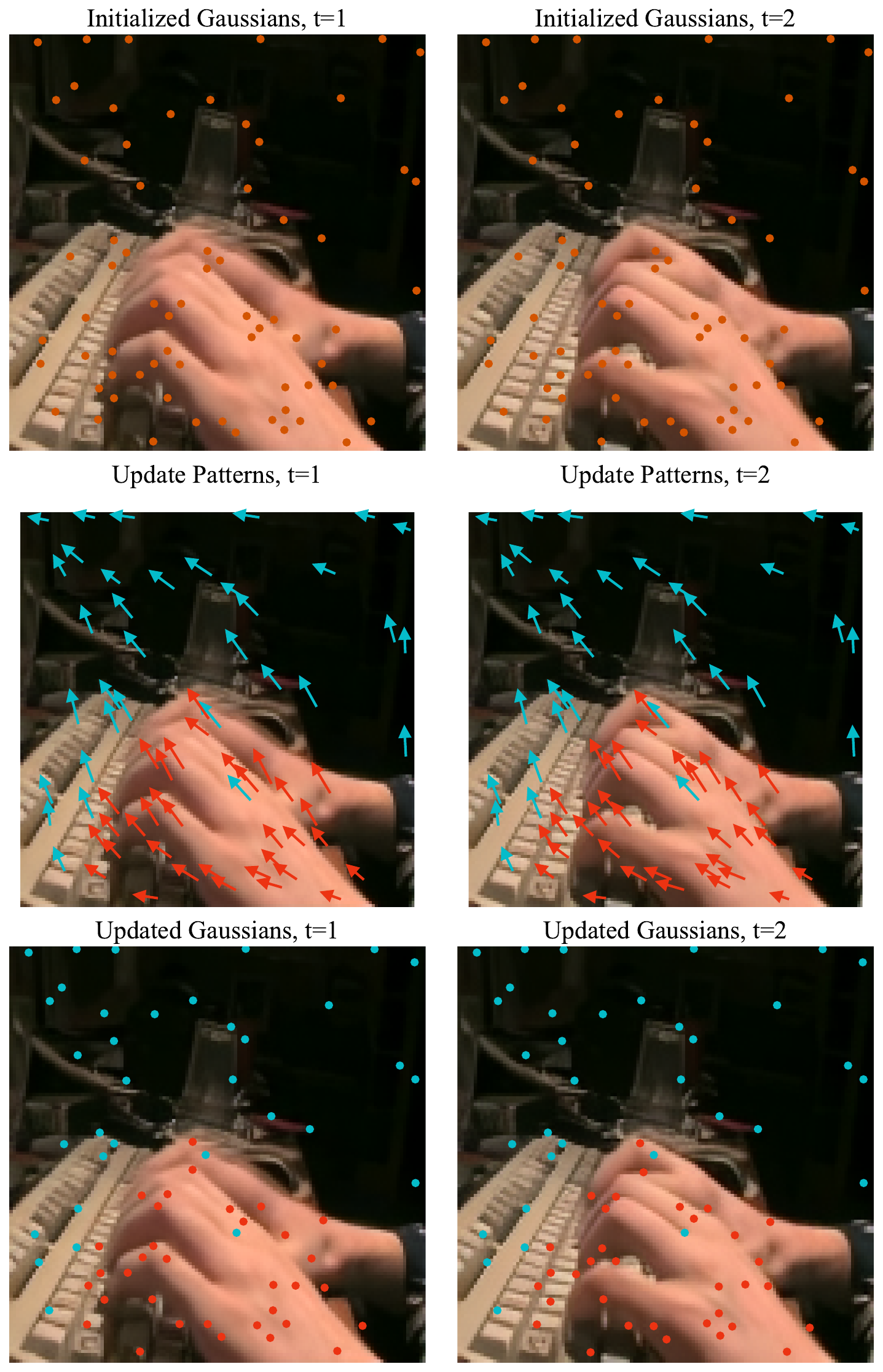}
  \label{fig:sup_motion}
\end{figure}

Since, during the STGE stage, each 2D Gaussian tensor is updated by adding the updated vector $\Delta \g_{t}^k$ (see \textit{Sec.~\ref{sec:gvt_dstf}}), the subsequent GSP procedure is expected to select $S$ Gaussians $\{\g_{t}^{k} \mid k=1:S\}$ whose updated vectors $\Delta \g_{t}^k$ remain highly similar across time-steps. In other words, after STGE, the difference between $\g_{t}^{k}$ and $\g_{t-1}^{k}$ for these Gaussians would be negligible (see \textit{Sec.~3.2.3}). Hence, we treat these 2D Gaussians as ``Static Gaussians" and retain only their representations at the first time step (\ie, $t=1$), forming $\G_{Static} = \{\g^{k,Static} \mid k=1:S\}$. We show this procedure in Fig.~\ref{fig:sup_motion}.

\subsubsection{Differentiable Binarization}
\label{sec:gvt_bin}
We follow the differentiable quantization strategy based on the Straight-Through Estimator (STE)~\cite{yinunderstanding} to generate a binary mask $\m$ from a learned mask vector $\g_{Mask}$ (see \textit{Sec. 3.2.3}). Specifically, we first apply a Sigmoid function to $\g_{Mask}$ to obtain a probability vector $\tilde{\m} \in \mathbb{R}^{K}$, where each element $\tilde{\m}_k \in (0,1)$.
We then binarize $\tilde{\m}$ into a hard mask $\m^{\text{Hard}} \in \mathbb{R}^{K}$, where $\m^{\text{Hard}}_k \in \{0, 1\}$ is set to $1$ if $\tilde{\m}_k > 0.5$, and $0$ otherwise.
To make this binarization differentiable, we apply the STE strategy using the formulation:
$\m = \text{SG}(\m^{Hard} - \tilde{\g}) + \tilde{\g}$,
where $\text{SG}(\cdot)$ denotes the stop-gradient operation, which prevents gradients from flowing through its argument. This approach enables the use of the hard binary mask $\m$ during the forward pass, while allowing gradients to propagate through the soft values $\tilde{\m}$ during the backward pass. We summarize the process using Algorithm~\ref{alg:bin}.

\begin{algorithm}[t!]
\caption{Differentiable Binarization with STE}
\begin{algorithmic}[1]
\Require Mask vector $\g_{Mask} \in \mathbb{R}^{K}$, threshold $\cH = 0.5$
\State $\tilde{\m} = \text{Sigmoid}(\g_{Mask})$ \Comment{Compute soft probabilities}
\For{$k = 1$ to $K$}
    \If{$\tilde{\m}_k >\cH$}
        \State $\m^{\text{Hard}}_k = 1$
    \Else
        \State $\m^{\text{Hard}}_k = 0$
    \EndIf
\EndFor
\State $\m = \text{Stop\_Gradient}(\m^{\text{Hard}} - \tilde{\m}) + \tilde{\m}$ \Comment{Apply STE}
\State \Return $\m$
\end{algorithmic}
\label{alg:bin}
\end{algorithm}

\subsection{Experimental Details}
\label{sec:supp_exp}
\subsubsection{Video Reconstruction}
\label{sec:supp_exp_vr}
For the video reconstruction results on UCF101 and K600 (\ie, Table1), we reproduce MAGVIT-v2 and LARP-L using the official open-source implementations~\cite{luo2024open, wanglarp}. Results for TATS and MAGVIT are reported from~\cite{luo2024open}, while those for SweetTok, MaskGIT and VQGAN are taken from~\cite{tan2024sweettok}. The results for OmniTok are obtained from its original paper~\cite{wang2024omnitokenizer}.  We provide the codebook sizes used by each video tokenizer in Table~\ref{tab:codebook_size}.

Regarding the DAVIS-2017-480P results (\ie, Table~2), we report the video representation performance of 4DGS, RoDynRF, Deformable Sprites, OmniMotion, CoDeF, and Splatter-a-Video using the results from~\cite{sun2024splatter}, and reproduce NeRV and HNeRV with their official open-source implementations~\cite{chen2021nerv, chen2023hnerv}. Here, in addition to LPIPS and SSIM, we also include PSNR-based benchmarks. 
It is worth noting that our optimization strategy is designed for perceptually realistic reconstruction (see \textit{Sec.~3.3}), rather than purely maximizing reconstruction fidelity as done by the competing methods in Table~\ref{tab:VR-2-1}. Nonetheless, GVT still achieves comparable PSNR performance.
To further demonstrate the versatility of our method, we additionally optimize a variant by omitting the generator loss $\cL_{Gen}$ from \textit{Eq.~10}. This variant, denoted as GVT (Reconstruction) in the last row of Table~\ref{tab:VR-2-1}, achieves 4.9 \% PSNR enhancement, indicating that our model can achieve even higher reconstruction quality with a slight modification of the optimization objective.

\subsubsection{Video Action Recognition}
\label{sec:supp_exp_var}
We use the official open-source implementation of VideoMAE~\cite{tong2022videomae} to conduct video action recognition experiments. Specifically, we input both the raw videos and the reconstructed videos from MAGVIT-v2 and our GVT into VideoMAE models trained on the UCF101 and K400 datasets. Recognition accuracy is evaluated on the UCF101 test set and the K400 validation set, following the protocol as in~\cite{tong2022videomae}.
To further validate video action recognition performance, we add results using the more recent action recognition methods UniFormerV2~\cite{li2022uniformerv2} and InternVideo2~\cite{wang2024internvideo2} at Table~\ref{tab:k400_cls}, where our GVT performs better than baseline, consistent with results in Table~\ref{tab:VAR}.

\begin{table}[!t]
\centering
\caption{Codebook size used by each video tokenizer, represented as \textbf{codebook length $\times$ latent dimension}.
$^*$MAGVIT-v2~\cite{yulanguage} adopts a lookup-free quantization strategy and does not use a conventional codebook. Following common practice~\cite{tan2024sweettok}, we denote its effective codebook size as 262144.}
\resizebox{0.6\linewidth}{!}{
\begin{tabular}{lc}
\noalign{\hrule height 1pt}
\textbf{Methods}         & \textbf{Codebook Size} \\
\noalign{\hrule height 1pt}
MaskGIT~\cite{chang2022maskgit}        & $1024 \times 256$          \\
VQGAN~\cite{esser2021taming}          & $1024 \times 256$          \\
TATS~\cite{ge2022long}           & $1024 \times 256$          \\
MAGVIT~\cite{yu2023magvit}         & $1024 \times 256$            \\
OmniTok~\cite{wang2024omnitokenizer}        & $8192 \times 8$            \\
LARP-L~\cite{wanglarp}         & $8192 \times 8$          \\
SweetTok~\cite{tan2024sweettok}  &  $26210 \times 256$ \\
MAGVIT-v2~\cite{yulanguage}        & $262144^*$       \\
\noalign{\hrule height 1pt}
GVT (UCF-101) & $4096 \times 8$         \\
GVT (K600)    & $4096 \times 8$        \\
\noalign{\hrule height 1pt}
\end{tabular}}
\label{tab:codebook_size}
\end{table}

\begin{table}[!t]
\centering
\caption{Video representation results in PSNR on DAVIS-2017-480P. Blue shading indicates \textbf{Top-3} methods: darkest ($1^{st}$), medium ($2^{nd}$), light ($3^
{rd}$).}
\resizebox{0.85\linewidth}{!}{
\begin{tabular}{l|cc}
\noalign{\hrule height 1pt}
\textbf{Methods} & \textbf{Fitting Time} & \textbf{PSNR$\uparrow$} \\
\noalign{\hrule height 1pt}
NeRV~\cite{chen2021nerv}                  & $\sim$45 mins        & 26.15                        \\
HNeRV~\cite{chen2023hnerv}                & $\sim$15 mins        & \toptwo{27.82}                        \\
4DGS~\cite{Wu_2024_CVPR}                  & $\sim$40 mins        & 18.12                         \\
RoDynRF~\cite{liu2023robust}              & \textgreater 24 hrs  & 24.79                         \\
Deformable Sprites~\cite{ye2022deformable}& $\sim$30 mins        & 22.83                         \\
OmniMotion~\cite{wang2023tracking}        & \textgreater 24 hrs  & 24.11                         \\
CoDeF~\cite{ouyang2024codef}              & $\sim$30 mins        & 26.17                \\
Splatter-a-Video~\cite{sun2024splatter}   & $\sim$30 mins        & \topone{28.63}     \\
MAGVIT-v2~\cite{yulanguage}                                     & Feed-Forward & 24.77            \\
\noalign{\hrule height 1pt}
GVT                                       & Feed-Forward & 25.12            \\
GVT (Reconstruction)                                      & Feed-Forward & \topthree{26.36}             \\
\noalign{\hrule height 1pt}
\end{tabular}}
\label{tab:VR-2-1}
\end{table}

\subsubsection{Video Compression}
\label{sec:supp_exp_vc}
For variable-bit-rate compression, we apply different quantization step sizes, $Q \in \{1\times10^{-3}, 2\times10^{-3}, 5\times10^{-3}\}$, to the latent representation $\bar{\mathbb{Z}}$ via the standard quantization equation
$
\tilde{\mathbb{Z}} = \mathrm{round}\left(\frac{\bar{\mathbb{Z}}}{Q}\right)\cdot Q
$.
We then employ the open-source CompressAI implementation~\cite{begaint2020compressai} of the hybrid entropy model proposed in~\cite{cheng2020learned} to estimate the bitrate $R(\tilde{\mathbb{Z}})$. With all other GVT modules frozen, the auxiliary entropy model is trained on the UCF101 and K600 training sets using the following rate-only objective:
\begin{equation}
\mathcal{L}_{\text{Compression}} = R(\tilde{\mathbb{Z}})
\label{eq:compression}
\end{equation}

\begin{table}[!t]
\centering
\caption{Additional Recognition Results on Kinetics-400.}
\resizebox{\linewidth}{!}{
\begin{tabular}{l|c|c|c|c}
\hline
\textit{Methods} &
\multicolumn{2}{c|}{UniFormerV2-B/16} &
\multicolumn{2}{c}{InternVideo2-B/14} \\
\cline{2-5}
& Top-1 (\%)$\uparrow$ & Top-5 (\%)$\uparrow$ & Top-1 (\%)$\uparrow$ & Top-5 (\%)$\uparrow$ \\
\hline
Raw Video & 85.80 & 97.05 & 88.39 & 98.04 \\
MAGVIT-v2 & 79.57 & 94.01 & 85.34 & 96.81 \\ \hline
GVT & 80.83 & 94.71 & 86.47 & 97.35 \\
\hline
\end{tabular}}
\label{tab:k400_cls}
\end{table}

We use VTM-23.4~\cite{vtm} and HM-18.0~\cite{HM} with the following:

\begin{itemize}
\item \textbf{HM}
\begin{verbatim}
TAppEncoder -c encoder_lowdelay_P_main.cfg [args]
\end{verbatim}

\item \textbf{VTM}
\begin{verbatim}
EncoderApp -c encoder_lowdelay_P_vtm.cfg [args]
\end{verbatim}

\end{itemize}

Both codecs use the following common line arguments ([args]):
\begin{verbatim}
    --InputFile={input_filename}
    --BitstreamFile={bitstream_filename}
    --ReconFile={reconstructed_filename}
    --InputBitDepth=8
    --OutputBitDepth=8
    --OutputBitDepthC=8
    --InputChromaFormat=420
    --IntraPeriod=-1
    --FramesToBeEncoded={frame_num}
    --SourceWidth={width}
    --SourceHeight={height}
    --FrameRate={frame_rate}
    --QP={quantization_parameter}
\end{verbatim}

The width, height, and number of frames are set to 128, 128, and 17, respectively, following the same input setting as our GVT and the baseline MAGVIT-v2 (see \textit{Sec.~4.1.2}).

\begin{algorithm}[t!]
\caption{Unconditional Video Generation Training}
\label{alg:uvg-train}
\begin{algorithmic}[1]
\Require clean latent $\mathbb{Z}$, auxiliary predictor $\Phi_\theta$
\State $u \sim \mathcal{N}(\mu,\sigma^2)$
\State $t = \mathrm{Sigmoid}(u)$ \Comment{sample $t$ from a logit-normal distribution}
\State $\epsilon \sim \mathcal{N}(0,\mathbf{I})$
\State $\mathbb{Z}_t = t\mathbb{Z} + (1-t)\epsilon$
\State $\mathbb{F}_t = (\mathbb{Z} - \mathbb{Z}_t)/(1-t)$ \Comment{target velocity field}
\State $\hat{\mathbb{Z}}_t = \Phi_\theta(\mathbb{Z}_t, t)$
\State $\hat{\mathbb{F}}_t = (\hat{\mathbb{Z}}_t - \mathbb{Z}_t)/(1-t)$ \Comment{predicted velocity field}
\State $\mathcal{L}_{Match} = \|\hat{\mathbb{F}}_t - \mathbb{F}_t\|_2^2$
\end{algorithmic}
\label{alg:uvg-train}
\end{algorithm}

\begin{algorithm}[t!]
\caption{Unconditional Video Generation Inference}
\label{alg:uvg-sample}
\begin{algorithmic}[1]
\Require auxiliary predictor $\Phi_\theta$, frozen GVT framework $\mathrm{GVT}(\cdot)$, time steps $\{t_n\}_{n=0}^{N}$ with $0=t_0<\cdots<t_N=1$
\State $\mathbb{Z}_{t_0} \sim \mathcal{N}(0,\mathbf{I})$
\For{$n=0$ to $N-1$} \Comment{progressively denoise the latent}
    \State $\hat{\mathbb{Z}}_{t_n} = \Phi_\theta(\mathbb{Z}_{t_n}, t_n)$
    \State $\hat{\mathbb{F}}_{t_n} = (\hat{\mathbb{Z}}_{t_n} - \mathbb{Z}_{t_n})/(1-t_n)$
    \State $\mathbb{Z}_{t_{n+1}} = \mathbb{Z}_{t_n} + (t_{n+1}-t_n)\hat{\mathbb{F}}_{t_n}$
\EndFor
\State \Return $\hat{\mathbb{V}} = \mathrm{GVT}(\mathbb{Z}_{t_N})$ \Comment{synthesize the video}
\end{algorithmic}
\label{alg:uvg-sample}
\end{algorithm}

\subsubsection{Unconditional Video Generation}
\label{sec:supp_exp_uvg}
We follow the JiT~\cite{li2025back} flow-matching scheme for unconditional video generation. During training, we sample $t$ from a logit-normal distribution, \ie, $u \sim \mathcal{N}(\mu,\sigma^2)$ and $t=\mathrm{Sigmoid}(u)$, and sample Gaussian noise $\epsilon \sim \mathcal{N}(0,\mathbf{I})$ to construct the noisy latent $\mathbb{Z}_t = t\mathbb{Z} + (1-t)\epsilon$. The target velocity is defined as $\mathbb{F}_t = (\mathbb{Z}-\mathbb{Z}_t)/(1-t)$. An auxiliary predictor $\Phi_\theta$ estimates the clean latent $\hat{\mathbb{Z}}_t$ and yields the predicted velocity $\hat{\mathbb{F}}_t = (\hat{\mathbb{Z}}_t-\mathbb{Z}_t)/(1-t)$. We optimize $\Phi_\theta$ with the $\ell_2$ velocity-matching loss as in Eq.~\ref{eq:match}, while keeping all GVT modules frozen. The whole training procedure is summarized in Algorithm~\ref{alg:uvg-train}. 

\begin{equation}
\mathcal{L}_{Match}=\|\hat{\mathbb{F}}_t-\mathbb{F}_t\|_2^2
\label{eq:match}
\end{equation}

During inference, generation starts from the Gaussian noise $\mathbb{Z}_{t_0=0}  \sim \mathcal{N}(0,\mathbf{I})$ and progressively integrates the latent trajectory from $t=0$ to $t=1$ using an Euler ordinary differential equation (ODE) solver, as summarized in Algorithm~\ref{alg:uvg-sample}. The denoised latent representation, $\mathbb{Z}_{t_N=1}$, is then passed through the frozen GVT framework, including STGE, GSP, vector quantization, reconstruction, and the decoder described in \textit{Sec.~3.2} and \textit{Sec.~3.3} of the main manuscript, thereby producing the final generated video.

In implementation, we use $\mu=-0.8$ and $\sigma=0.8$ to sample $t$ during the training procedure, and adopt a simple Euler solver with $N=50$ sampling steps during the inference procedure. The remaining implementation details follow JiT~\cite{li2025back}.


\end{document}